\documentclass{llncs}
\usepackage[utf8]{inputenc}

\usepackage{color}
\usepackage{todonotes}
\usepackage[normalem]{ulem}

\usepackage{graphicx}

\title{
Early Classification of Time Series is Meaningful
}

\author{Youssef Achenchabe\inst{1,2}, Alexis Bondu\inst{2},\\ Antoine Cornuéjols\inst{1}, and Vincent Lemaire\inst{2}}

\institute{UMR MIA-Paris, AgroParisTech, INRAe, 
              Universit\'e Paris-Saclay, 
              France \\
 \and
Orange Labs, 44 Avenue de la r\'epublique, Ch\^atillon, France\\
}

\begin{document}

\maketitle

\begin{abstract}
Many approaches have been proposed for early classification of time series in light of its significance in a wide range of applications including healthcare, transportation and finance. However, recently a preprint saved on Arxiv claims that all research done for almost 20 years now on the Early Classification of Time Series is useless, or, at the very least, ill-oriented because severely lacking a strong ground. In this paper, we answer in detail the main issues and misunderstandings raised, and propose directions to further expand the fields of application of early classification of time series.

\keywords{early classification  \and cost estimation \and online decision making}
\end{abstract}

\section{Introduction}

An  increasing  number  of  applications  require  to  recognize  the  class of  an  incoming  time  series  as  quickly  as  possible  without  unduly  compromising the  accuracy  of  the  prediction \cite{teinemaa2018alarm,alipour2019machine,russwurm2019early,fahrenkrog2019fire,sharma2020novel,loyola2017learning,dachraoui2013early}. For example, in emergency wards of hospitals \cite{mathukia2015modified}, in control rooms of national or international electrical power grids, in government councils assessing emergency situations, in all kinds of contexts, it is essential to make timely decisions in absence of complete knowledge of the true outcome  (e.g. should the patient undergo a risky surgical operation?). 
The issue facing the decision makers is that, usually, the longer the decision is delayed, the clearer is the likely outcome (e.g. the critical or not critical state of the patient) but, also, the higher the cost that will be incurred if only because earlier decisions allow one to be better prepared. How to optimize online the tradeoff between the earliness and the accuracy of the decision is the object of the Early Classification of Time Series problem (ECTS).

To the best of our knowledge, \cite{alonso2004boosting} is the earliest paper explicitly mentioning ``classification when only part of the series are presented to the classifier'', and the main thrust of it is to show how the boosting method can be employed to the classification of incomplete time series.

The ECTS community has produced a large diversity of models that can predict the class of incoming subsequence, after only seeing a fraction of the data. This community has published his work on diverse venues in prestigious conferences or journals \cite{Schfer2020TEASEREA,dblp2485352,ruwurm2019endtoend,Ghalwash2014,Mori2016ReliableEC,Parrish2013ClassifyingWC,Dachraoui2015,mori2015early,mori2017early,Wang2016EarlinessAwareDC,tavenard2016cost,martinez2018deep,anderson2012early,Yao2019DTECDT,8765556,He2019ConfidencebasedEC,9066213,achenchabe2021MLj,Yan2020ExtractingDF}. Thus, step after step, this community is doing progress towards solving this interesting problem.

However, a recent preprint \cite{wu2021early} essentially claims that all research done for almost 20 years now on the Early Classification of Time Series is useless, or, at the very least, ill-oriented because severely lacking a strong ground. 

\section{Arguments against ECTS}

Once \cite{wu2021early} dissected and the line of reasoning made apparent, in fact, the critic laid down in the paper boils down to three arguments:

\begin{enumerate}
   \item The paper exhibits cases where early classification of time series is doomed to lead to numerous false positives and false negatives, so as to be effectively useless.  The paper claims that this must be the general case. The fact that no real application is to be found in the scientific litterature, so it is said, is a testimony of the fact that ECTS is a false problem.
   
   \item The research conducted on ECTS has mostly, if not exclusively, used the UCR repository, and hence has produced results that are misleading. There would be three reasons for this. \textit{First}, the UCR datasets contain only time series of the same length and starting at the same time. This is hardly what can be expected in the real world. \textit{Second}, most of the time series in these datasets can be classified observing only their starting subsequence. This would credit ECTS systems with classification performances that are unearned. And \textit{third}, the authors claim that most ECTS system require a z-normalization of the data, which can only be obtained afterwards and thus removes all relevance to making early classification in the first place.
   
   \item Since, it is easy to obtain similar performances as ECTS systems with just fumbling a little bit with the data and using classical time series classification algorithms, it is not worth resorting to the overhead of ECTS systems. 
   
\end{enumerate}

Let us consider each argument in turn. 

\section{Discussion}

\subsection{ECTS is a false and flawed problem}

It is interesting that Evolution has not entirely wiped out this argument. Every creature on earth who said to itself 
``\textit{Ah, I see an animal in that tree that looks like a feline, a leopard maybe or a saber-toothed tiger, but I'm not sure. It's partly hidden in the foliage. Funny, these movements, maybe it's getting ready to jump in my direction. It's not completely clear though. Would it want to eat me? Maybe I'm wrong. Let's wait to find out more. Early danger classification is a false problem, a recent paper stated'}' must be dead and without descendant. 
In the wild, animals behave appropriately by fleeing at the right moment in front of a predator, by finding a compromise between the earliness and the accuracy of their decision. To illustrate this compromise, an animal should not flee at the slightest noise because it would waste its energy most of the time, and it should flee before its decision is completely confident, i.e. when the predator is fully visible and pouncing on it. Optimizing such a compromise is the essence of ECTS.

In \cite{wu2021early} the authors of the paper make the point that it is actually a very risky enterprise to classify time series on the basis of a starting subsequence since, so they argue, it is so common that such subsequences are followed by different continuations. But here, the argument is flawed on two accounts at least.


First, the particular examples presented by the authors do not constitute evidence validating their point of view in its generality, but only a confirmation bias in order to support author's prior beliefs.
In particular, the metaphor drawing a parallel between the ECTS and word recognition in textual data {\textit{distorts the problem}}\footnote{In the case of ECTS, labels are assigned to an {{\bf entire}} time series, which has a beginning and an end. Moreover, the class is not necessarily defined by a {{\bf local pattern}} to be found.}, which  does not help the reader to correctly understand the ECTS. The example from Levicitus 2:1 is reductive and far-fetched. Certainly, there are many examples of words that share subparts, but ECTS is not reduced to that.
Definitely, such examples do not constitute a proof that ``the space of possible domains where ECTS could be meaningly applied is vanishingly small''  as authors claim. And even if that was the case, remember that general relativity was long believed to have no practical applications until the GPS appeared. Now, we could simply not do without general relativity. But this is not the major shortcoming of the critic addressed by the authors. 

The second shortcoming is that ECTS systems are designed to learn from training data that potentially includes ambiguous examples, but this point does not seem to be addressed in \cite{wu2021early}. 
When such ambiguous examples in the input signal, these systems just wait to have more information before making their decision. So, if indeed the syllables ``cat'' on one hand, and ``dog'' on the other hand, are shared by different words that appear statistically often, like ``catalog'' and ``cattle'' or ``dogmatic'' and ``doggedly'', then these systems will simply not base their decisions on these syllables alone. That's all there is to it.

In a way, we can liken the work on ECTS to what is studied in control theory \cite{bennett1993history,bellman1964control}. Control theory is concerned as well with online decision making. This is the case with anti-aircraft guns that must point ahead of the plane in order to hit it. And the farther away is the plane from the gun, the more ahead of the plane should the pointing be. Here also, the plane could try to evade the antigun shots by changing its trajectory. The very same course  can be followed by different ones. Does that render control theory useless? Notice that a whole lot of standard rules in control theory use as well integration of measurements over some time interval, thus taking stock of the past to take decisions, exactly as ECTS does. Early classification of time series goes one step further though, and tries to learn second order knowledge. The idea is to learn \textit{when} there will be enough information to decide in such a way as to optimize both the expectation of the misclassification cost and the cost associated with delaying the decision. 

At this point, it is interesting to discuss what the authors \cite{wu2021early} believe is a counter example of the usefulness of ECTS. They dispute the usefulness of a early warning system in hospitals which would be able to detect early on abnormal ECG, because, so they say, these systems would only bring a gain of at most seconds in detection when surgical operations often take days to be planned. This just points out that, in this context, the cost of delaying decision is in fact very low. And, in consequence, there is no need to hurry the classification of time series. In other contexts, such as planning the start-up of thermal power plants from consumption measurements to face a peak of consumption on the day, gaining just one half an hour can be critical. So, this is not the ECTS systems that are at fault. Their added value depends both on the misclassification costs and the cost associated with delaying the decision. If the later is low, then the best course of action can be to wait for all information to be available before committing to a decision.

The thing is though, that it is not always clear that all the information is available. Everyday, we have to take decisions from incomplete information. For instance, I decide to sit on this brown contraption of which I see only two legs because my past history of encounters with such things leads me to predict that it is a chair. Should I bother to control that all four legs are indeed present each time I plan to seat on what appears to be a chair, but may be not?
We make such inferences all the time from incomplete information with the risk of course to make faulty interpretations and predictions. Should we stop doing this\footnote{In \cite{wu2021early}, as of the date 03-02-2021 \textit{``the community should stop publishing on this topic''}.}, as the authors of the paper seem to imply and demand? Would that only be possible? Would this even be viable?

As Machine Learning will gain in prominence and be employed in all facets of life, learning under budget constraints will become quite a significant topic. Reducing the number of measurements can be crucial in some applications, and this is what gives so much value to innovative methods like compressive sensing \cite{candes2006stable}.  ECTS belongs to this family of methods that seek to be parsimonious in the measurements they need. 

\subsection{ The research has been ill-oriented because of the almost exclusive use of UCR time series}

We think claim that this line of criticism is circumstantial and does not put into question the validity of the interest for ECTS. One reason for using the UCR datasets is simply that everyone does this, and that it helps making comparisons between the various proposed approaches. 
%


Another reason is that, in fact, applications where ECTS is useful exist \cite{teinemaa2018alarm,alipour2019machine,russwurm2019early,fahrenkrog2019fire,sharma2020novel,loyola2017learning,dachraoui2013early} but they have still to realize their full potential. Companies are becoming more and more aware of the problem and are now starting to integrate ECTS into their projects. 

Let us now discuss the objections made in turn. The \textit{first} one is that the UCR datasets contain only time series of the same length and starting at the same time. This indeed facilitates the task of devising and learning classifiers that make predictions given some part of the time series. But it must be said that the point of ECTS is {\bf not} to devise new classifiers, or classifiers that are particularly effective. The problem instead is to take existing classifiers and look at ways to define a criterion that combines the misclassification costs and delay cost, and then to devise ways to optimize it. A fair comparison between ECTS methods must therefore use the very same classifiers. Now, if datasets exist where time series are not of the same length and same starting point, this would indeed be interesting, but comparisons of ECTS methods would still require that the same classifiers be used, here classifiers able to deal with these other scenario.

\textit{Another} objection is that numerous of the oldest methods of classification of time series 
imply the use of distances such as the Euclidian distance or DTW which are quite sensitive to the fact that the data has been z-normalized\footnote{~The use of such data is explained by the fact that the first research works on ECTS first exploited methods based on distances or similarity measures, like K-Nearest Neighbor, etc. For this reason, some of the UCR data donors have provided pre-normalized datasets.}. Because z-normalization is based on the knowledge of the whole time series, this means that the start of a time series ``leaks'' information about its remaining part, and this in turn biases the performances that are reported for some ECTS systems. This interesting remark, while casting doubts on the absolute performance reported for some ECTS systems, does not however condemn the whole field of ECTS since many methods {do not require} 
this normalization {(e.g. SR \cite{mori2017early}, Economy-$\gamma$ \cite{achenchabe2021MLj} and all other approaches that do not use a similarity measure)}. In addition, more and more datasets are now put in UCR and in other data bases which are not z-normalized a priori. It is recommended that the community exclude the few z-normalized datasets from the reference benchmark to compare ECTS approaches.

\textit{Finally}, \cite{wu2021early} observes that most of the time series in these datasets can be classified observing only their starting subsequence. This would allow ECTS systems to be credited with unearned classification performance. 
Again, the point of the research in ECTS is not to show that it is possible to label time series given some starting (small) subsequences. It is to show that the combined misclassification cost and delay cost can be lowered using ECTS. This is what must be evaluated. Believing that what counts is the proportion of the time series needed to label it,  
is a misunderstanding of 
the ECTS problem. 

\subsection{It is not worth resorting to the overhead of ECTS systems}

This is an interesting question. If ECTS claims to take into account the delay cost in the classification problem, which is not considered classically, then it would only seem fair to take also into account the cost of using the machinery needed to perform ECTS. So, what are the overheads implied with ECTS systems?

Let us take a very recent system, which explicitly optimizes the combined misclassification cost and delay cost in a non myopic manner, that is by predicting at each time step what is the expected best instant to make a decision, and which exhibits state of the art performances \cite{achenchabe2021MLj}. 

\smallskip
There, the cost of using this system is entailed by: 
\begin{itemize}
    \item training $N$ classifiers, when the length of the time series is divided into $N$ intervals. Thus, a classifier $h_i (i \in \{1, \ldots, N\})$ is trained to label time series when the first $i$ time intervals are provided. 
    \item computing groups of time series. In \cite{achenchabe2021MLj}, it is shown that this can be done in several ways, supervised or unsupervised. 
\end{itemize}
These steps are performed during training. They therefore not entail cost during recognition. 

When in use, at the time interval $i$, the cost comes from the evaluation of the future combined cost for all remaining $N - i$ time steps. This evaluation is fast, involving assigning the incoming time series to a group of time series, and then estimating the expected misclassification cost in this group and the delay cost, which is a function of time. If the time seems optimal to take a decision, then the incoming time series is labeled according to the current classifier. 

The overhead incurred during recognition is therefore quite low. And all the process is automated, while what \cite{wu2021early} describe as ``this took common sense and a few minutes of low-code exploration of the data'' in order to be able to label time series with some prefix subsequence does not guarantee the quality of the result in any quantitative way, and certainly not the optimization of the combined misclassification cost and delay cost. 

\section{Sum up}

Our artificial classification systems, as ourselves, must  perform inferences from incomplete and ambiguous input data all the time. The problem is to act as well as possible given this limited information. ECTS goes one step further. It tries to estimate when misclassification costs combined with a delay cost can be optimized. Note that the delay cost can include the cost of acquiring one more measurement. This is thus a very significant line of research. Giving examples where early classification of time series is risky does not provide evidence of the uselessness of ECTS. These systems would detect the high risk of misclassification given the sparse information provided and would balance it against the cost of delaying decision. If the delay cost is low enough, they would wait for more measurements. Otherwise, they would take the (high) risk of miclassification because it is still less than the cost of delaying decision. 

That the research carried out so far has not lead to applications is to our opinion a mistaken critic {\cite{teinemaa2018alarm,alipour2019machine,russwurm2019early,fahrenkrog2019fire,sharma2020novel,loyola2017learning,dachraoui2013early}}. 
{Real applications are: i) either hidden from the academic world because of their sensitive nature; ii) or under development, as this is a relatively new problem whose awareness is recent.}
Furthermore, it is fortunate that entire fields of research have not been terminated in the past because of the apparent lack of potential applications. We would still be relying on Chappe's telegraph to exchange information at distance, since Maxwell's research on electromagnetism was deemed to be without any concrete applications. And we would not have computers because Turing's early research was completely abstract. 

Finally, the current stance on using datasets with special properties (i.e. the UCR dataset), while being worth of warnings, does not condemn the field to a very limited set of applications. It was necessary to test the various methods on a shared test bed. ECTS is now moving toward new questions, for example early detection of anomalies in data streams. 
Recent papers propose solid formalizations of the ECTS problem 
\cite{achenchabe2021MLj,mori2019early,russwurm2019end,hartvigsen2019adaptive}, which are not considered in the state of the art presented in \cite{wu2021early} that is biased towards oldest works related to ECTS.


\section{Conclusion and open challenges}

Early Classification of Time Series is definitely a meaningful problem for several reasons. First, numerous applications require ECTS, and all organisms on Earth are confronted to this problem. Saying that this is too hard a problem and therefore that research efforts aimed at it are worthless amounts to renouncing to do science. Besides, the main criticism raised by Wu et al. \cite{wu2021early} about the high risk of misclassification in ECTS is not new. This is the issue ever present in supervised learning. The problem is to learn what is relevant, or discriminant, in the input signal and what is not or what is ambiguous. The novelty in ECTS is that, at test time, a tradeoff between the leisure to take as much information as possible before making a classification and a cost for waiting has to be solved. Directly confronting the problem has been shown to increase the performance as compared to approaches that do not. 

ECTS raises interesting problems that can be matched with problems in control theory. As in control theory, decisions have to be made online and must take into account what is the most likely future given what is known about the past, but, in addition, ECTS seeks to anticipate the best moment to take a decision under delay costs. 

There are also other open
challenges and future topics of research in early classification as for example: 
(i) extend ECTS to the classification of portions of data streams (e.g.: to tackle predictive maintenance, or anomaly detection usecases);  
(ii) extend ECTS to revocable decisions allowing to correct a bad decision when the recent measurements suggest that the predicted class is wrong (e.g.: useful for monitoring systems); 
(iii) extend ECTS to \textit{unsupervised} or \textit{weakly supervised} learning tasks (e.g. outliers or anomaly detection).
These three points open interesting research perspectives that would allow to redefine and extend the ECTS problem, and therefore to further expand its fields of application.
To conclude, we consider that ECTS is a well-defined problem thanks to the years of sustained work in this field \cite{achenchabe2021MLj,mori2019early,russwurm2019end,hartvigsen2019adaptive}.  We encourage the scientific community to work on ECTS, as we are convinced that it is a promising area.

\bibliographystyle{splncs04}
\bibliography{references}
\end{document}